\documentclass[letterpaper]{article}
\usepackage{iccc}
\usepackage{algorithmic}
\usepackage[linesnumbered,ruled,vlined]{algorithm2e}
\usepackage{multirow}
\usepackage{graphicx}

\usepackage{times}
\usepackage{helvet}
\usepackage{courier}
\title{Conceptual Expansion Neural Architecture Search (CENAS)}
\author{
Mohan Singamsetti$^{1,2}$, Anmol Mahajan$^{1,2}$, and Matthew Guzdial$^{1,2}$\\
$^1$Computing Science Department, University of Alberta\\
$^2$Alberta Machine Intelligence Institute (Amii)\\
\{singamse, mahajan, guzdial\}@ualberta.ca,
}
\begin{document} 
\maketitle
\begin{abstract}
\begin{quote}
Architecture search optimizes the structure of a neural network for some task instead of relying on manual authoring. 
However, it is slow, as each potential architecture is typically trained from scratch. 
In this paper we present an approach called Conceptual Expansion Neural Architecture Search (CENAS) that combines a sample-efficient, computational creativity-inspired transfer learning approach with neural architecture search. 
This approach finds models faster than naive architecture search via transferring existing weights to approximate the parameters of the new model.
It outperforms standard transfer learning by allowing for the addition of features instead of only modifying existing features. 
We demonstrate that our approach outperforms standard neural architecture search and transfer learning methods in terms of efficiency, performance, and parameter counts on a variety of transfer learning tasks.
\end{quote}
\end{abstract}

\section{Introduction}

Deep learning is the study of deep neural networks (DNNs), which are a type of function approximators. 
DNNs have achieved remarkable success in various challenging applications such as image classification, image generation, and natural language processing \cite{szegedy2015going}.
Modern deep learning approaches perform well when researchers train large models, often with at least millions of parameters, on large amounts of data \cite{halevy2009unreasonable,brown2020language}. 
However, this leads to two problems. 
First, the size of these models limits where they can be applied and who can afford to train them.
Second, there are application domains in which we do not have sufficient training data, and therefore where we cannot currently apply these approaches.

The limitations of large pre-authored architectures have been addressed using Neural Architecture Search (NAS). 
In this approach, the model architecture is optimized along with the model weights.
NAS has outperformed manually designed architectures in some tasks such as object detection \cite{zoph2018learning} and image classification \cite{zoph2018learning}.
However, NAS is a time consuming and computationally expensive process \cite{li2019random} since it requires training many potential architectures from scratch.

Approaches exist to transfer the knowledge of models trained on large source datasets to tasks with smaller target datasets, including transfer learning \cite{lampert2009learning}, domain adaptation \cite{daume2009frustratingly}, few-shot learning \cite{fei2006one} and zero-shot learning \cite{xian2017zero}. 
However, these approaches all require re-training the models \cite{levy2012teaching}, or require manually authoring or learning secondary features \cite{xian2017zero} to handle new cases or to adapt to a new domain. 
In addition, these approaches generally assume fixed architectures set by a human expert according to the target task.
This is a limiting factor, as the transfer learning process can be forced to adapt knowledge rather than retain it due to the fixed size. 
For example, when attempting to adapt a source network trained to recognize cats to a target network to recognize dogs some features would be better to retain (e.g. fur), while others would be better to replace (e.g. cat eyes).  

Intuitively, if we could combine NAS and transfer learning we could end up with an approach that could find more efficient models more quickly, even for cases with less data.
NAS could benefit from transfer learning as weights can be transferred from existing networks, speeding up the process of training novel architectures. 
Transfer learning could benefit from NAS as it can allow for the addition of new features instead of only the modification of existing features learned from a source dataset.
We believe this has the potential to lead to smaller and more accurate models that can be trained more quickly, and therefore represents a valuable open problem.
Beyond recent domain adaptation and architecture search work \cite{li2020network}, this combination of neural architecture search and meta learning is a largely unexplored area of research.
One of the reasons for this may be that transfer learning often represents a much faster and simpler optimization problem than training from scratch. 
In comparison to this, naive architecture search represents an unbounded search problem.

Combinational creativity \cite{boden2004creative}, also called conceptual combination \cite{gagne1997influence}, is a cognitive process in which old knowledge is combined to produce new knowledge.
This is a general process in human cognition \cite{gagne1997influence,fauconnier2001conceptual}, an efficient means of representing new concepts with existing knowledge. 
Attempts have been made to approximate this cognitive process computationally, most famously conceptual blending, to the point where any attempt to approximate combinational creativity with computation is called blending \cite{fauconnier2001conceptual}.
However, these earlier combinational creativity approaches are generally limited to hand-authored inputs and curated knowledge bases. 

Our problem in this paper directly relates to combinational creativity \cite{boden2004creative}.
In combinational creativity, existing knowledge is recombined to created new knowledge. 
As a cognitive process, it is unlikely that our human brains replicate the existing knowledge in order to produce a recombination, as this would be inefficient and slow. 
Instead, evidence suggests this is a quick, compact process \cite{gagne1997influence}. 
We therefore argue that simultaneous neural architecture search and transfer learning is a reasonable computational metaphor for combinational creativity in neural networks.

To address our problem we employ a representation that allows for sample-efficient transfer learning called Conceptual Expansion \cite{guzdial2018combinets,banerjee2021combinets,guzdial2021conceptual}.
In this transfer learning approach the reuse of a source model's knowledge is modeled as a combinational creativity problem.
With conceptual expansion we can approximate the weights of a target model as a combination of weights from a source model. 
This allows for a much faster optimization of model weights, therefore speeding up architecture search. 
We call our approach \emph{Conceptual Expansion Neural Architecture Search (CENAS)}. 
In a number of image classification domains we demonstrate how CENAS outperforms standard architecture search, transfer learning, and naive architecture search with transfer learning. 
This work contributes this novel approach, experimental results that demonstrate that it outperforms existing approaches to meta learning and architecture search \cite{li2020network}, and earlier applications of conceptual expansion to deep neural networks \cite{guzdial2018combinets,banerjee2021combinets}.

\section{Related Work}

In this section we overview the two most related areas of prior work: architecture search and transfer learning.

\subsection{Architecture Search} 

Architecture search attempts to automatically determine the optimal neural network architecture for a particular problem.
The approach dates back to the 1980's, when evolutionary optimization approaches were proposed to find both the architectures and weights of a neural network \cite{miller1989designing}. 
As the name implies, the problem is typically represented as a search problem, where some initial architecture or population of architectures are optimized to find the best architecture in a given search space \cite{xie2017genetic}.  
We employ evolutionary search in this work as it has been shown to still be the best or equivalent optimization method for neural architecture search \cite{real2019regularized}.
Many optimization strategies have been used to explore the space of the possible architectures \cite{ZophL16}.
However, architecture search methods still struggle to find the same results as architectures hand-authored by human experts. 

\subsection{Transfer Learning}

Transfer learning of deep neural networks (DNN) refers to the transfer of knowledge from a DNN trained to solve one source problem to a DNN for designed to solve a related target problem.
A wide range of prior approaches exist for the transfer of knowledge in neural networks such as domain adaptation and one or zero-shot learning \cite{fei2006one,xian2017zero}. 
These kinds of approaches often require additional features to guide the transfer of knowledge, which can be hand-authored or learned from a secondary dataset \cite{ganin2016domain}. 
Our approach does not use any additional hand-authored or machine-learned features.

Domain adaptation focuses on transferring the knowledge gained from one or more labelled domains to an unlabelled target domain. 
Multi Source Domain Adaptation (MSDA) takes training data collected from multiple sources and applies that to a single unlabelled target \cite{peng18}. 
Neural Architecture Search for Domain Adaptation (NASDA) is a recent approach focused on deriving the best architecture for a specific domain adaptation task by leveraging differentiable neural architecture search \cite{li2020network}. 
Our approach also uses a unique representation similar to NASDA, but it is a representation that approximates novel class features in the target domain as combinations of source domain features.

Combinational creativity \cite{boden2004creative} or conceptual combination \cite{gagne1997influence} represents the ability of humans to combine existing knowledge to produce new knowledge. 
Computational implementations of combinational creativity can be understood as a specialized case of transfer learning, focused on re-representing existing knowledge to approximate new or unseen concepts. 
There have been many combinational creativity approaches with conceptual blending being the most popular \cite{fauconnier2001conceptual,guzdial2018combinatorial}. 
However, the majority of these existing approaches can only take hand-authored symbolic data as input.
Guzdial and Riedl introduced combinets \cite{guzdial2018combinets}, the application of combinational creativity to deep neural networks via a representation they called Conceptual Expansion, which is designed to work with messy, machine-learned knowledge.
We build directly upon this work, but extend it to a more general approach that better leverages the available, existing data to do simultaneous transfer learning and architecture search.

\section{System Overview}

\begin{figure}[tb]
    \centering
  \includegraphics[width=3in]{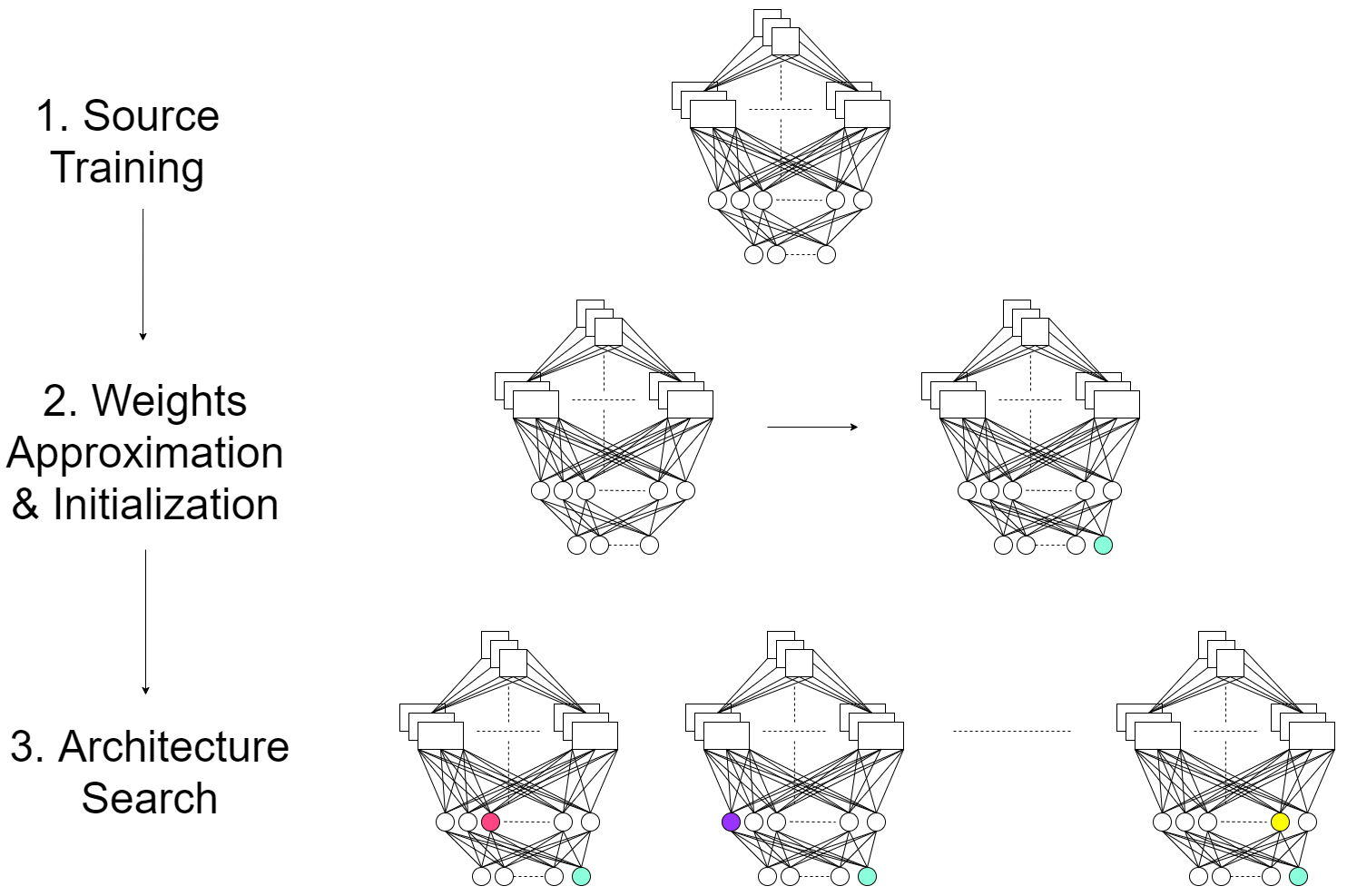}
  \caption{The three step CENAS Process. First we train a model in our source domain. Second, we approximate the weights of the initial model in the target domain as a conceptual expansion of the source model's weights. Finally, we run architecture search, changing the architecture and approximating the new weights of the architecture simultaneously.}
  \label{fig:CENAS_framework}
\end{figure}

In this section we present our Conceptual Expansion Neural Architecture Search (CENAS) approach. 
This approach is focused on domain transfer problems: where we have distinct source and target datasets, and where the goal is to adapt knowledge from the source domain to the target domain. 
We define CENAS as a three-step process.
First, we train a model on the source dataset. 
Second, we approximate the weights of the connections of an initial target model as a conceptual expansion: a combination of weights from the source model.
Third, we run our architecture search process on target model by updating our approximations of existing connections and approximating the weights of any additional connections in the same manner. 
This process is visualized in Figure \ref{fig:CENAS_framework}.

\subsection{Conceptual Expansion}

We begin by formally defining Conceptual Expansion (CE). 
CE is a way of representing knowledge as a combination of existing knowledge, in our case neural network weights.
If we wished to represent a particular weight $w$ as a combination of existing weights with CE we would use Equation~\ref{eq1}:
\begin{equation} \label{eq1}
  \mathrm{CE^w}(F,A) = a_1*f_1+a_2*f_2+... +a_n*f_n
\end{equation}

\noindent
where the $F$ = {$f_{1}$, $f_{2}$, ......., $f_{n}$} represent a set of existing weights, $A$={$a_{1}$, $a_{2}$, ....., $a_{n}$} are alpha filter matrices that undergo pairwise multiplication with their paired existing weight matrix. 
$A$ act as instructions for how to transform the weight for the combination.
Notably, the same weight value can appear multiple times in $F$ with different $a$ values.
CE can be understood as analogous to the crossover function in evolutionary search \cite{whitley1994genetic}, both rely on the intuition that combinations of high-quality knowledge are more likely to lead to new high-quality knowledge. 
Similarly, crossover can be understood as another example of combinational creativity, the general human cognitive process for reusing old knowledge \cite{gagne1997influence,fauconnier2001conceptual}.
CE is designed to be a simple but extendable way to represent combinations of neural network weights, in order to more clearly study combinational creativity in DNNs.
In the original CE paper \cite{guzdial2018combinets}, Guzdial and Riedl employed greedy search to optimize the $f$ and $a$ values, finding that this approach outperformed backpropagation-based transfer learning approaches for low sample sizes. The values in $a$ matrix range [-2, 2].
However, the greedy optimization failed to outperform existing methods for larger sample sizes of data. 

\subsection{Source Training}

In this paper we focus on the image classification domain, employing several common image classification datasets as our source and target datasets. 
We employ CifarNet as our initial source model throughout this paper \cite{krizhevsky2009learning}, as it represents a well-understood and compact initial architecture. 
CifarNet has two convolutional layers each with max pooling and two fully connected layers. We implemented it unchanged from the original description \cite{krizhevsky2009learning}. The convolutional layers apply the convolution operation to the input in order to extract features. Max Pooling is a convolution process where it down-samples the feature representation by taking only the maximum values.
Our first step is to train CifarNet on the source dataset.

\subsection{Weights Approximation}

The second step of our approach approximates the weights on an initial target model.
Given that we focus on image classification in this paper, we only need to approximate novel weights for the final classification layer in this step.
To approximate these weights we take our available training data for each target class and pass it through the model trained on the source dataset. 
This gives us a distribution over the $n$ source classes for each target class (e.g. In a target domain ``fox'' class images might be classified as the source domain class ``dog'' 65\% of the time and the remaining 35\% of the time as ``cat'').
We normalize these values with a softmax function, which gives us our initial $a$ values. 
Softmax is an activation function which maps the output in the range [0, 1] and also maps each output in such a way that the total sum is 1.
Every one of the non-zero values output from the softmax function is paired with its source $w$ value and combined to approximate the initial target class $w$ value as in Equation \ref{eq1} (e.g. the source domain class cat and dog weights are associated with their alpha values to approximate the initial target domain class fox classification neuron weight). 
For all other layers in the model we initialize their $a$ value as a matrix of ones of the appropriate shape for that layer's weights. 
Thus the initial model is represented as a conceptual expansion that is equivalent to the source model except for the final layer.

\SetKwInput{KwInput}{Input}                
\SetKwInput{KwOutput}{Output}    

\begin{algorithm}[tb]
\SetAlgoLined
\KwInput{An architecture $A$, the population size $pop\_size$, maximal generations $gen$, the $source$ dataset, and the $target$ dataset.}
\KwOutput{Best performing architecture.}
 $A$ $\leftarrow$ train $A$ on $source$\;
 $A$ $\leftarrow$ Re-represent $A$ using CE and $target$\;
 $pop$ = \small\{$A$ \small\}\;
 
 \While{$|pop|$ $<$  $pop\_size$}{
 $network$ $\leftarrow$ Mutation($A$)\;
 $pop$.append($network$)\;
 }
 i $\leftarrow$ 0\;
 \While{i $<$ gen}{
$pop$ $\leftarrow$ Crossover($pop$)\;
$pop$ $\leftarrow$ Mutation($pop$, $mutationRate$)\;
$fitness\_pop$ = Fitness($pop$)\;
$pop$ $\leftarrow$ Reduce($pop$, $fitness\_pop$)\;
i $\leftarrow$ i + 1\;
 }
architecture = best\_model(pop)\;

Return architecture\;
 \caption{CENAS Workflow}
 \label{Algo_workflow}
\end{algorithm}

\subsection{Architecture Search}

\begin{algorithm}
\SetAlgoLined

\KwInput{An architecture $Net$, target domain data $D_{target}$ with $N$ classes of image dataset, $target\_classes$ are list of classes in target domain}
\KwOutput{Fitness scores of $Net$.}
score $\leftarrow$ 0;\

\For{class in $target\_classes$}{
score $\leftarrow$ score + $Net$.accuracy($D_{target}[class]$)}\;
Return score/N;

\caption{Fitness Score}
\label{Fitness_Eval}

\end{algorithm}

For the architecture search step of CENAS we use an evolutionary optimization process or genetic algorithm, given their history and consistent performance in NAS tasks \cite{real2019regularized}.
We chose this as greedy optimization of CE struggles to exit the local optima near the model output from the weight approximation step \cite{guzdial2018combinets,banerjee2021combinets}.
Evolutionary optimization requires that we initialize a population of points, define mutation and crossover operators, and a fitness function. 
We represent this whole process in Algorithm 1, with lines one and two representing the first and second steps of CENAS described above, and the remaining lines devoted to this final step.  
We initialize a population of fixed size based on the output of the second step, running our mutation function $pop\_size$-1 times to produce each population member.
From there we iterate through the standard evolutionary search steps, running our crossover function to double our population size, mutating the population members, evaluating the new models on the target domain training data, and reducing back to our original population size. 
The mutation and crossover functions act directly on the model architecture and weights; we describe them in detail below.
We explored several fitness functions, but found that taking the average accuracy over the target domain training data gave the best performance as represented in Algorithm \ref{Fitness_Eval}.


\subsection{Mutation and Crossover}

For our crossover function we use a simple single-point crossover. We take two models and target a random CNN layer from each for the split point. 
We then take the first half of one model and the second half of the other.

We employ a total of seven different mutation operators. Our first four operators are typical for architecture search applications: 
\begin{itemize}
\item The first mutation operation adds a new convolutional layer to the architecture at a random position before fully connected layers and after the first convolutional layer.
We randomly choose 32 filters or 64 filters and a kernel shape of 3x3 or 5x5, we use all the other parameters from the original CifarNet's convolutional layers. 
\item The second mutation operation deletes a random convolutional layer in the network besides the first one to maintain the fixed input size.
\item The third mutation operation adds additional filters to an existing convolutional layer. We randomly choose a convolutional layer in the network except for the first layer and add a random filter count of 2, 4, 8, 16, or 32.
\item The fourth mutation operation deletes filters in a random convolutional layer of the network; we delete a random filter count of 2, 4, 8, 16, or 32 filters.
\end{itemize}

\noindent
The remaining three mutation operations help in manipulating the network weights directly by modifying the $a$ and $f$ values associated with each weight.
\begin{itemize}
\item The fifth mutation operation multiplies the $a$ of a random $f$ of a random layer by a scalar in the range [-2,2].
\item The sixth mutation operation replaces a $f$ value of a random layer with a randomly selected $f$ value. 
\item The seventh mutation operation adds a random $a$ and $f$ to a random position in a random layer (e.g. adding $a_{n+1}$*$f_{n+1}$ to $a_1$*$f_1$+$a_2$*$f_2$... $a_n$*$f_n$).
\end{itemize}

\section{Experiments}

We focus on convolutional neural networks (CNNs) for our initial exploration of CENAS, as they were used in prior combinets \cite{guzdial2018combinets,banerjee2021combinets}.
We explore this through two major types of experiments in this paper. 
First, we measure the performance of CENAS on several tasks that have been used in prior domain transfer and architecture search work, alongside several baselines \cite{li2020network}.
This is meant to present evidence for our claims of the value of CENAS to transfer and architecture search tasks.
Second, we present a series of low data $n \rightarrow n+1$ tasks using the CIFAR-10 \cite{krizhevsky2009learning} dataset. 
This is meant to present evidence for our claims around how CENAS can operate even for low data problems, which relates to how humans can employ combinational creativity with few examples \cite{gagne1997influence}.
In addition, this second evaluation allows us to directly compare to the original CE with DNNs work \cite{guzdial2018combinets}.

For all tasks we make use of CifarNet as our base architecture.
We train CifarNet for 100 epochs on our source dataset.
For CENAS we then use the the training dataset as described above to guide the search over models for 100 generations with a population of size 10.
We chose these low values to demonstrate the effectiveness of the approach with minimal computation, and as part of our investigation as to whether this can be an appropriate metaphor for combinational creativity in neural networks.
We take the final 10 members of the final generation and train them for 30 epochs on our target dataset in a standard supervised learning paradigm using RMSProp. 
In all cases we use a batch size of 32 and a learning rate of 0.0001.
We used Keras for implementation and non-CENAS training of our deep neural networks.

We have four baselines. 
The first two are variations of CENAS. 
R-CENAS employs a random walk instead of a genetic algorithm.
It uses the same seven mutation functions from above and chooses one at random for every architecture. 
We run the random walk for 100 steps and output the top five best models according to target training accuracy.
G-CENAS makes use of greedy optimization instead of a genetic algorithm.
We try 10 random mutation functions at every step as a neighbor function and choose the best across the neighbors and current model according to training accuracy. 
We run for 100 iterations and take the final five models as our output. 
Afterwards, we take this output and train it for 30 epochs using the target training data.

The next two methods represent how one might naively attempt to solve this problem using more standard methods.
The first of these is a simple neural architecture search implementation (NAS). 
For this implementation we used the same fitness and crossover functions from our CENAS implementation, but only the first four mutation functions, making it a more standard NAS implementation.
After mutation and crossover functions we instantiate the new model and train on the available target training data for 30 epochs. 
If this naive NAS outperforms CENAS it would indicate that our approach to transferring existing features via recombination is actively detrimental.
Finally, we include a naive combination of neural architecture search and transfer learning (NAS-T). 
This is similar to our naive NAS implementation but we transfer existing weights from the parent models during crossover and copy over the weights from the most similar weight or filter for our mutation functions.
We then train on the target training data for 12 epochs, as we found that any more training led to overfitting.
If NAS-T outperforms CENAS that indicates that our more complex representation has no benefit over simply finetuning the existing weights, and is therefore no better as a metaphor for human combinational creativity.
For all of the NAS approaches (NAS, NAS-T, and CENAS) that rely on a genetic algorithm we report the results for the top five members of the final generation according to training accuracy.

All the experiments are carried out using the cloud computing resources of Compute Canada, which uses 32 cores 4 x NVIDIA V100 Volta (32G HBM2 memory). 
We employ a consistent random seed across all experiments.

\begin{table*}[tb]
\caption{Domain Transfer Tasks Average Accuracy and Parameter Count Results}
\small
\centering
\begin{tabular}{|c|c|c|c|c|c|c|c|c|}
\hline
\multirow{2}{*}{}    & \multicolumn{2}{c|}{MNIST -$>$ USPS}            & \multicolumn{2}{c|}{USPS -$>$ MNIST}            & \multicolumn{2}{c|}{STL -$>$ CIFAR-10}           & \multicolumn{2}{c|}{CIFAR-10 -$>$ STL}           \\ \cline{2-9} 
                           & test acc       & model para     & test acc       & model para    & test acc      & model para      & test acc       & model para      \\ \hline

R-CENAS & 74.9 $\pm$ 15 & 1.8M $\pm$ 9.9k & 99.3 $\pm$ 0.0 & 2.1M $\pm$ 846k         & 11.2 $\pm$ 0.3          & 3.6M $\pm$ 2.3M         & 11.1 $\pm$ 0.1         & 2.1M $\pm$ 1.5M         \\ \hline
G-CENAS               & 90.0 $\pm$ 8.3          & 903k $\pm$ 12.2k         & 99.3 $\pm$ 0.1          & 1.7M $\pm$ 615k         & 81.7 $\pm$ 0.5          & 2.3M $\pm$ 1.1k         & 74.9 $\pm$ 0.5          & 2.2M $\pm$ 451k          \\ \hline
NAS                & 83.5 $\pm$ 12         & 1.5M $\pm$ 55.6k        & 99.3 $\pm$ 0.1          & 1.3M $\pm$ 539k         & 75.7 $\pm$ 1.2          & 4.2M $\pm$ 59k           & 59.8 $\pm$ 2.0          & 2.1M $\pm$ 213k          \\ \hline
NAS-T & \textbf{95.5 $\pm$ 2.4} & 1.6M $\pm$ 13.8k        & 99.2 $\pm$ 0.1          & 1.5M $\pm$ 13.6k        & 78.8 $\pm$ 0.1          & 3.7M $\pm$ 131k          & 61.7 $\pm$ 1.8          & 3.3M $\pm$ 157k          \\ \hline
CENAS & 89.94 $\pm$ 12 & \textbf{579k $\pm$ 534k} & \textbf{99.4 $\pm$ 0.1} & \textbf{920k $\pm$ 511k} & \textbf{82.5 $\pm$ 0.6} & \textbf{2.2M $\pm$ 170k} & \textbf{77.7 $\pm$ 2.8} & \textbf{1.9M $\pm$ 295k} \\ \hline
\end{tabular}
\label{table:domain_transfer_results}
\end{table*}

\begin{table}[tb]
\caption{Domain Transfer Tasks Average Computation Time in GPU Hours/Days}
\begin{center}
\begin{tabular}{ |l|c| } 
 \hline
 Approach & Average Hours (GPU Hours)\\ 
 \hline
 R-CENAS & 8H 36M \\ 
 \hline
 G-CENAS & 6H 55M \\ 
 \hline
 NAS & 2D 9H \\ 
 \hline
 NAS-T & 2D 6H \\ 
 \hline
 CENAS & 7H 23M \\ 
 \hline
 
 \hline
\end{tabular}
\end{center}
\label{table:domain_transfer_time}
\end{table}

\subsection{Domain Transfer Experiments}

\subsubsection{Setup} 

We make use of four tasks and datasets inspired from prior domain adaptation work \cite{hoffman2018cycada,li2020network}.
The four datasets are MNIST, USPS, STL-10 \cite{coates2011analysis}, and CIFAR-10 \cite{krizhevsky2009learning}, which are all well-understood image classification datasets.
Each of our tasks involves using one of the datasets as a source and another as a target, making our four tasks: MNIST$\rightarrow$USPS, USPS$\rightarrow$MNIST, STL-10$\rightarrow$CIFAR-10, and CIFAR-10$\rightarrow$STL-10. 
Given that CifarNet was designed for CIFAR-10 we modify the other datasets to have the same 32x32 input size, but do not otherwise process them, unlike in prior domain adaptation work where certain classes are removed \cite{hoffman2018cycada,li2020network}.

\subsubsection{Results}

We present the average test accuracy and standard deviation for each target dataset using the given test splits, and the average parameter count of the output models for each approach in Table \ref{table:domain_transfer_results}.
As a comparison point, the default CifarNet has 597K parameters.
Overall, CENAS outperformed the baselines on three of the four tasks. 
The only task it struggled on was the MNIST$\rightarrow$USPS task, which seems to be a difficult domain transfer task given prior results \cite{li2020network}. 
We anticipate this is due to transferring from a monochrome to an RGB colour domain. 
The NAS approaches that involved backpropagation to a greater extent were able to better adapt to this new domain. 
However, while our final CENAS models were roughly 5\% less accurate they were also three times smaller.
Of particular interest are the USPS$\rightarrow$MNIST results, which outperform even supervised domain transfer approaches reported in prior work \cite{hoffman2018cycada}.

While we do not include the results in the table or describe them as baselines, the very first model of the first generation for NAS and NAS-T represent training CifarNet on the target domain from scratch and finetuning CifarNet trained on the source domain respectively. 
Our finetuned CifarNet test accuracy (36.07, 99.37, 78.49, 77.64 on the four tasks) outperformed several baselines, but didn't outperform CENAS. 
Thus, CENAS seems to be a better metaphor for the human ability to adapt to new knowledge by reusing existing knowledge than standard transfer learning.
Comparatively, CifarNet trained only from scratch on the target domain with no adaptation or transfer outperformed CENAS in all but the third task (96.41, 99.52, 81.60, 78.51).
This is still a positive result overall as NAS architectures tend to struggle to even equal the performance of hand-authored architectures \cite{saxena2016convolutional}.
Put another way, CENAS seems to be the best approach for automatically adapting to new knowledge, but not for learning knowledge directly.

In terms of average parameter count, CENAS is a clear winner, with hundreds of thousands of parameters fewer than the closest approach. 
This may seem unintuitive, but a similar effect is seen with network pruning \cite{liu2018rethinking} where reducing the number of weights can be beneficial as it leads to more general models. 
Notably, the fitness function did not bias the output towards smaller models, only accuracy. 
The models in the final iteration are also generally larger than the initial architecture.
Instead, compactness arose as a secondary effect of pursuing accuracy and the combinational representation.
These compact representations parallel prior cognitive science results in terms of the representative power of combinations \cite{gagne1997influence,fauconnier2001conceptual}.
This provides some evidence that this may be true in deep neural networks as well. 
Of particular note is the relative size to the relative performance of these models, with some of these final models performing comparatively to models up to 3-4 times their size \cite{kabir2020spinalnet}.
Further, while CENAS is weakly supervised up to its final generation, prior unsupervised domain adaptation methods for these tasks far exceeded these parameter counts \cite{li2020network}.

We present the average computation time in GPU hours in Table \ref{table:domain_transfer_time}. 
R-CENAS was slightly slower than CENAS as the mutation functions occurred at every step, instead of only with some probability. 
While G-CENAS was on average somewhat faster due to early convergence, it's clear from the average parameters of its output models that it was biased towards adding features. 
The big difference is between the approaches that were strongly supervised, that re-trained at every generation: NAS and NAS-T. 
While NAS-T was somewhat faster as it trained for less time to adapt the existing features, our CENAS approaches were, on average, three times faster than these methods.

\begin{figure*}[t]
\centering
  \includegraphics[width=7in]{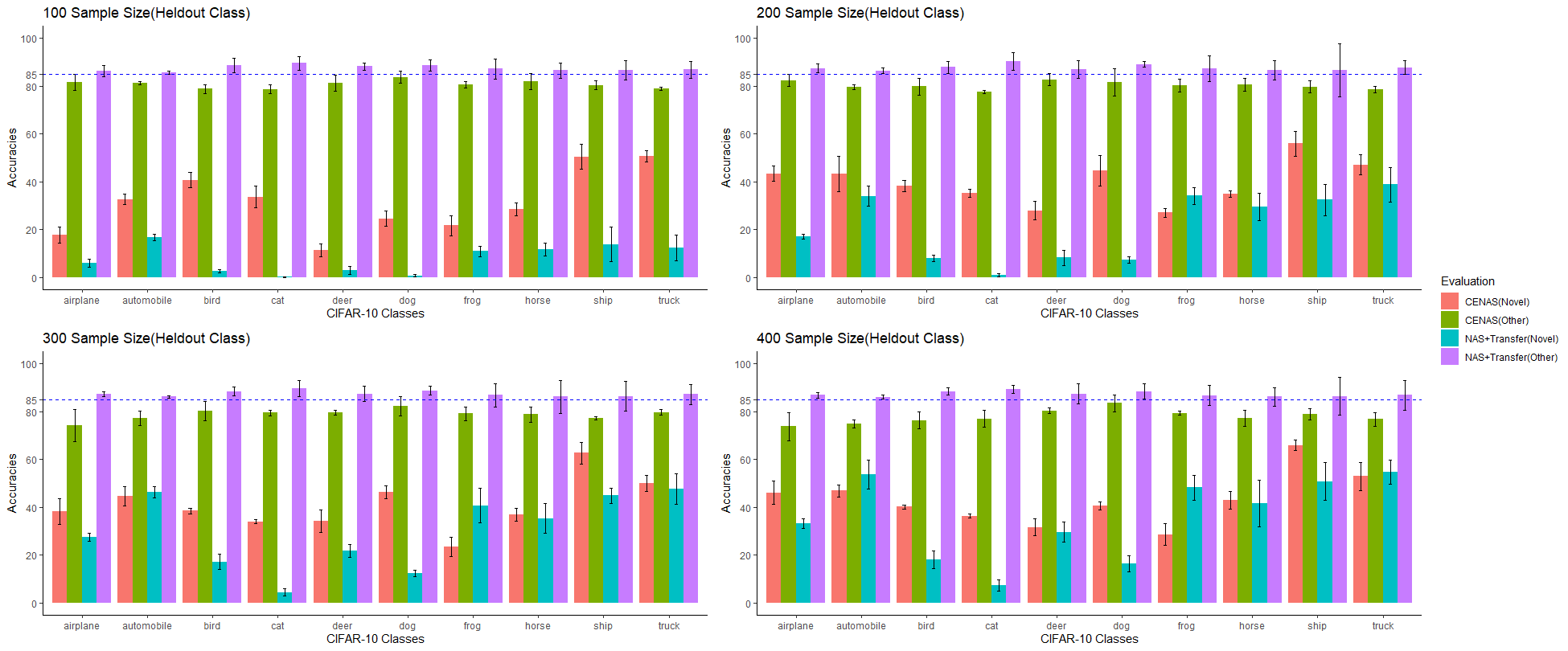}
  \caption{Evaluation results for 100-400 samples of the novel class.}
  \label{fig:All_sample_evaluation}
\end{figure*}

\begin{figure}[tb]
\centering
  \includegraphics[width=3in]{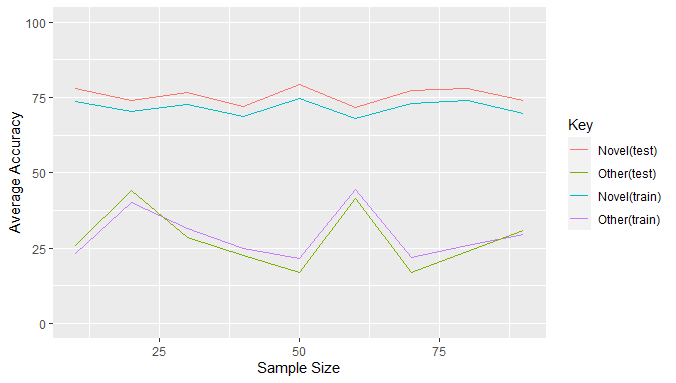}
  \caption{The average test and training accuracy for the CENAS approach for the 10 to 90 sample size cases.}
  \label{fig:10To90}
\end{figure}

\begin{figure*}[t]
\centering
  \includegraphics[width=6in]{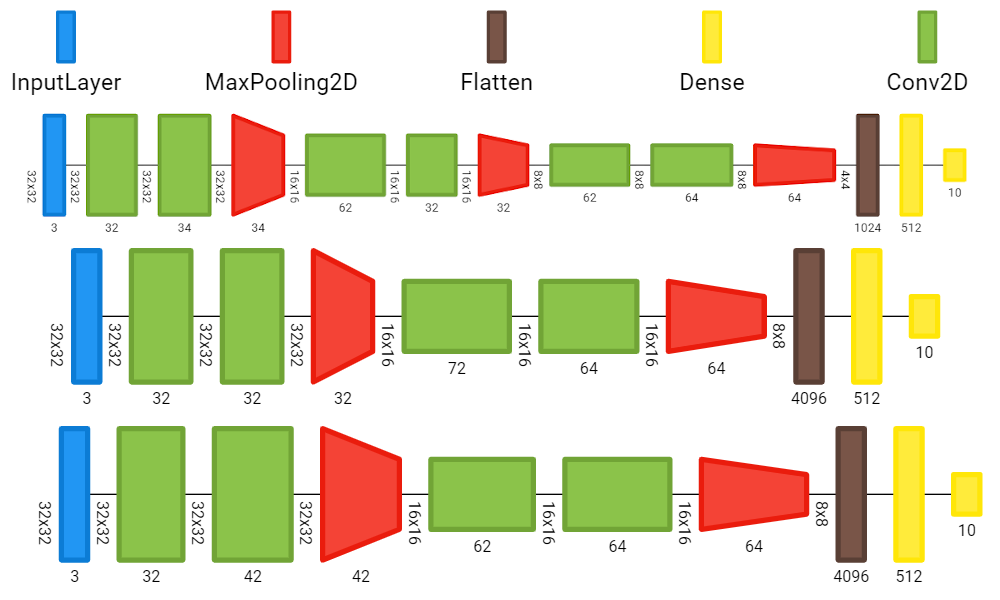}
  \caption{Example of three different architectures from the final CENAS generation for the held out airplane class.}
  \label{fig:All_architectures}
\end{figure*}

\subsection{$n \rightarrow n+1$ Experiments}

\subsubsection{Setup}
The results from our first experiments indicate CENAS can produce output domain transfer models that perform well, take less time to compute, and are more compact than similar performing models.
These features compare favorably to the results of studies on human combinational creativity \cite{gagne1997influence}.
However, these results came about using thousands of datapoints available in our target datasets, which is not similar to the human cognitive process. 
In the original paper by Guzdial and Riedl \cite{guzdial2018combinets}, they focused on how conceptual expansion could allow for low-data transfer in terms of the $n \rightarrow n+1$ problem.
To determine whether CENAS is still capable of handling this type of problem we ran a series of $n \rightarrow n+1$ experiments using the CIFAR-10 dataset. 
For each experiment we trained CifarNet on only 9 of the 10 available classes. 
We constructed a target dataset that included the training data of these 9 ($n$) classes and between 10 to 500 samples of the held-out ($n+1$) or Novel class. 
To ensure our results were reproducible, we made use of the first $X$ training instances in each experiment where $X$ is the sample size of the training images of the Novel class.
Notably we did not make use of backpropagation for the CENAS approach for this experiment, as it was not used in the original paper.

\subsubsection{Results}
We visualize the results for the $X$=100-400 cases per novel class in Figure \ref{fig:All_sample_evaluation}.
We only include the NAS-T baseline due to space constraints, and as NAS performed equivalently to NAS-T for these sample sizes.
We also note that our experiments included a transfer-only (no architecture search) baseline, which performed significantly worse.
The dotted line across all the graphs represents 85\%, which is the reported CIFAR-10 test accuracy for CifarNet, though we only observed values closer to 80\% when training on all available data \cite{krizhevsky2009learning}. 
From the Figure \ref{fig:All_sample_evaluation} it's clear that CENAS outperforms the baselines when it comes to the held out or novel class at lower sample sizes. 
We found that NAS-T and NAS began to perform equivalently or better than CENAS without backpropagation at and above the $X$=400 case. 
However, they had the same drawbacks as above in terms of model size and computation time.

We found that the accuracy on the held out class dropped to nearly 0 for all of our baselines for the $X<$100 case.
We visualize $X$=10-90 sample sizes for CENAS separately in Figure \ref{fig:10To90}.
Each line indicates the average across the novel classes of the novel ($n+1$) and other ($n$) classes.
Interestingly, CENAS retains better than chance accuracy on the held out class ($>$10\%) all the way down to 10 samples.
These results mirror the earlier results with greedy optimization and without architecture search \cite{guzdial2018combinets}, and outperform follow-up work using other optimization methods \cite{banerjee2021combinets}.

Interestingly, CENAS' performance on the held out class does not correlate to the sample size. 
We hypothesize that instead of training data size, a secondary feature of the held-out class training set is more important: the extent to which it reflects the true variance of the class in question. 
We also anticipate that this is closer to human combinational creativity, though we are unaware of prior work that investigates this.
If this is true, it could lead to even stronger results with tailored datasets.
We hope to study this in future work.


\section{Limitations and Future Work}

In this paper we focus solely on image classification domains for simultaneous architecture search and transfer learning. While our results do reflect the appropriateness of CENAS to these domains, even in cases with small amounts of available training data, this is still a major limiting factor. 
We plan to explore CENAS in sequence and generative modeling domains and with different types of data like audio and text in the future, in order to ascertain of these results hold.

Our current CENAS implementation relies on a fairly straightforward evolutionary search process. 
However, given that this search space is unbounded, it is unlikely that we have discovered the true global maxima. 
Simply increasing the number of generations or the population size is unlikely to solve this problem. 
We are currently exploring alternative strategies for more fully exploring this space, including ways to estimate the probable value of different operators in certain locations of the space or enforcing diversity with approaches like MAP-Elites \cite{mouret2015illuminating}.

\section{Conclusions}

In this paper we argue for exploring the problem of simultaneous architecture search and transfer learning as it relates to combinational creativity, and introduce an approach we call Conceptual Expansion Neural Architecture Search (CENAS). 
This approach relies on a neural representation of combinational creativity, the ability of humans to combine existing knowledge to produce novel solutions. 
We compare our approach to a set of baselines on several experiments using well-known image classification domains.
From this, we identify CENAS as a fast and sample-efficient method that produces high-quality and compact models.

\section{Ethics Statement}

There are a variety of potential concerns for any approach that seeks to lower resource requirements to apply deep neural networks. 
Specifically, there are ways in which bad actors could theoretically use an approach like CENAS to, for example, derive an image classifier for a particular person faster and with fewer images of said individual.
While we did not explore it in this work, prior work with Conceptual Expansion considered the generative case along with the discriminative case \cite{guzdial2018combinets}. 
Thus, there is a possibility that one could employ CENAS to more easily produce things like ``deep fakes''. 
However, concerns of this nature are premature, given that right now CENAS has only been evaluated in one domain. 
To combat this potential in future work, we intend to explore how CENAS models can be identified from their output. 

\section{Acknowledgments}

We gratefully acknowledge Vellore Institute of Technology -Semester Abroad Program for supporting this research. We also acknowledge the support of the Natural Sciences and Engineering Research Council of Canada (NSERC).






\bibliographystyle{iccc}
\bibliography{iccc}

\end{document}